\documentclass{article}

\usepackage[final, nonatbib]{unireps_2023}

\usepackage{amsmath,amsfonts,bm}

\def\Figref#1{Figure~\ref{#1}}

\def\eqref#1{equation~\ref{#1}}

\def\1{\bm{1}}

\def\mE{{\bm{E}}}

\def\mU{{\bm{U}}}

\DeclareMathAlphabet{\mathsfit}{\encodingdefault}{\sfdefault}{m}{sl}
\SetMathAlphabet{\mathsfit}{bold}{\encodingdefault}{\sfdefault}{bx}{n}

\def\gU{{\mathcal{U}}}

\def\sI{{\mathbb{I}}}

\usepackage[utf8]{inputenc} 
\usepackage[T1]{fontenc}    
\usepackage{hyperref}       
\usepackage{url}            
\usepackage{booktabs}       
\usepackage{amsfonts}       
\usepackage{nicefrac}       
\usepackage{microtype}      
\usepackage{xcolor}         
\usepackage{graphicx}
\usepackage{subcaption}

\title{Role Taxonomy of Units in Deep Neural Networks}

\author{%
Yang Zhao\qquad \qquad \quad  Hao Zhang\qquad  \qquad \quad Xiuyuan Hu \\
  Department of Electronic Engineering\\
  Tsinghua University\\
  \texttt{\{zhao-yang, haozhang\}@tsinghua.edu.cn, huxy22@mails.tsinghua.edu.cn} \\
}

\begin{document}

\maketitle

\begin{abstract}
    Identifying the role of network units in deep neural networks (DNNs) is critical in many aspects including giving understandings on the mechanisms of DNNs and building basic connections between deep learning and neuroscience. However, there remains unclear on which roles the units in DNNs with different generalization ability could present. To this end, we give role taxonomy of units in DNNs via introducing the retrieval-of-function test, where units are categorized into four types in terms of their functional preference on separately the training set and testing set. We show that ratios of the four categories are highly associated with the generalization ability of DNNs from two distinct perspectives, based on which we give signs of DNNs with well generalization.
\end{abstract}

\section{Introduction}

Deep neural networks (DNNs) are computing system modeled in an artificial way mimicking the biological neural networks, which have shown extraordinary ability in tackling real-world problems such as vision tasks \cite{DBLP:conf/cvpr/HeZRS16,redmon2016you}, natural language processing \cite{devlin2018bert,vaswani2017attention} and so on. Being the primary element in both the biological and artificial neural networks, units are activated to perceive the underlying features for a given task. Similar to the biological neural networks, units in DNNs could learn to present strong specialization and high function preference to specific subtasks in the task.

However, for each basic unit, its content is quite simple and abstract. It seems impossible to decode the meaning of units from a theoretical view. So research is mostly carried out from discussing what the functionality may present by each unit \cite{zeiler2014visualizing, dosovitskiy2016inverting, bau2017network, bau2020understanding}. Generally, the functionality of a unit describes its role in the neural network at task level, and is formed by the natural selection of neural unit groups during task learning. At the same time, each unit has different functional characteristics. When faced with different tasks, due to the distributed working mode in the connectionist neural network models, units will respond to the task according to their own functional characteristics, and thus play distinct roles \cite{hunsaker2013operation, denny2014hippocampal}.


In this paper, we are going to give a kind of role taxonomy of network units in DNNs. Before the taxonomy, we introduce a useful tool, called retrieval-of-function test, for identifying the group of units that could present the highest contribution on the given subtask when being activated jointly. In general, this group of active units may perform differently on the training set and the testing set. Based on this fact, we find that the units at a layer could be divided into four categories in terms of their function on the two sets: core units, overfitted units, generalizing units and confusing units. By illustrating with network models with different generalization ability, we show that the ratios of the four types of units could be closely associated with the model generalization. Finally, we give further discussions from the perspective of role shares of individual units to show the different behaviors they may present in different network models.

\section{Related works}



Focusing on the investigation of network units, one line of research is to indicate the working status of units to the given task with certain attribute. Typical attributes include L1-norm of units \cite{luo2017thinet} and the Average Percentage of Zeros (APoZ) of units when using Relu activation function \cite{hu2016network}. In addition, \cite{morcos2018importance} introduce the class selectivity from neuroscience to investigate the selectivity over classes for a specific unit, on the basis of calculating the mean feature maps. \cite{zhao2021quantitative} propose a kind of topological entropy for indicating the chaotic degree of feature pattern for individual units. Studying on the indication of the unit working status could tell quantitative understanding of the contribution of a unit to a given task. But, it is still unclear what role the units could play in this task.

Another related research trace is to find and visualize the elements in the input that are important to the units. In this way, the critical element found in the input could be considered as the features that the unit could perceive, which somehow unveils the functionality of a unit. Typical works \cite{zeiler2014visualizing, zhou2014object, mahendran2015understanding, simonyan2013deep} search for the image patches that maximize the score of a given unit. Alternatively, instead of searching, \cite{nguyen2016synthesizing, dosovitskiy2016inverting} generate related images considered the most important for a unit. Bau {et al.} measure the degree of conceptual alignment between units and human-visual concepts (Mountain, River, etc.) predicted by another vision segmentation model in their works \cite{bau2017network, zhou2018interpreting, bau2020understanding}. However, \cite{wang2020high} argue that functionalities of units in CNNs may be beyond the recognition to human vision.

\section{Identifying active network units via retrieval-of-function test}

In neuroscience, to identify the function of units, it is critical to locate the collaborative unit group that are highly active for given a specific task, which could be detected via the functional magnetic resonance imaging. Correspondingly, consider tackling a typical classification task with DNNs, and our investigation begins with identifying this group of units in DNNs that are highly active to a specific class in this task.

\begin{figure*}[http]
\begin{center}
\centering
\includegraphics[width=1\columnwidth]{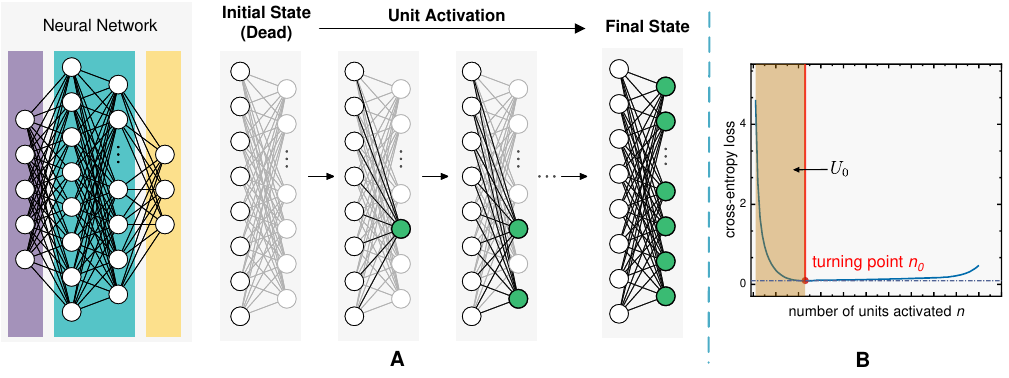}
\caption{Illustration of retrieval-of-function test (A) and its ROF curve in regards to the cross-entropy loss (B).}
\label{fig:cumulative unit activation}
\end{center}
\vskip -0.2in
\end{figure*}

\paragraph{Retrieval-of-function Test} To identify these highly active units, we will employ the retrieval-of-function (ROF) test, as illustrated in Fig.\ref{fig:cumulative unit activation}A. The core idea of ROF test is to identify the group of units that could recover the full perception of tasks when the networks stay in the "dead" state.

Basically, the ROF test could be summarized into two steps:
\begin{itemize}
   \item Firstly, deactivate all the units within the layer to force the DNN to stay at the "dead" state, where the network gives no response on any input.
   \item Secondly, iteratively activate the units according to an ordered list $\mU_{l} = \{\gU_i\}_{i = 0}^N$ individually. Typically, the ordered list could be ranked by quantities indicating the activation strength. Here, L1-norm will be used as the target quantity to give a descending rank of units in the ROF test. In this way, units are activated sequentially from $\gU_0$ to $\gU_N$ in the initial dead network.
\end{itemize}

During the process of activation, the evolution of performance with respect to the number of units activated (notated as $n$) could be recorded as the characterization to indicate the growing effect on network performance when activating the unit. We would call this recorded curve the ROF curve and notate it as $\mE(n)$. Fig.\ref{fig:cumulative unit activation}B shows a typical evolution of $\mE(n)$, where the cross-entropy loss is applied for recoding the performance of the DNN.

\paragraph{Turning point} Before performing unit activation, as deactivating all the units in the layer, the DNN is unable to perceive any valid feature. With the ordered list in a descending rank, units with strong activation would be activated at the beginning of the evolution process. The performance would experience a continuously rapid increase since the neural network gradually recover its function on perceiving the feature in this class. Notably, the cross-entropy loss would reach at the minimum after activating only a small part of units, after which it would begin dropping steadily until all the units have been activated. We mark the minimum point as the turning point of $\mE(n)$,
\begin{equation}
n_{0}= \mathop{\arg\min}_n \ \mE(n)
\end{equation}
And correspondingly, the group of units that have been activated could be identified as follows,
\begin{equation}
\mU_0 = \{\mathcal{U}_i\}_{i = 0}^{n_{0}}
\end{equation}
It should be pointed out that without loss of generality, $\{\gU_i\}_{i = 0}^N$ denotes the ordered list ranked by the descending order of the activation strength. $\mU_0$ represents the minimum group of units being activated jointly that could stimulate the most performance on the given class. We call $\mU_0$ the active unit set, and call its unit the highly active unit.


\section{Role taxonomy of units in DNNs}

In practice, to check the generalization ability, units would be trained on the training set $\mathcal{D}^{(train)}$ and tested for generalization on a held-out testing set $\mathcal{D}^{(test)}$. Notably, all the samples in this class should be drawn from the same distribution, sharing the same common features. For ideal networks, units should be able to perceive the common intrinsic features in the class. In this way, units could form substantial representations for unseen samples. This implies that units that are highly active on $\mathcal{D}^{(train)}$ are expected to perform in the same way on $\mathcal{D}^{(test)}$.

\begin{figure*}[http]
    \begin{center}
    \centerline{\includegraphics[width=0.8\columnwidth]{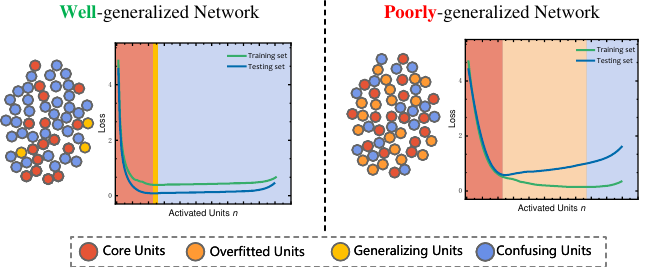}}
    \caption{Illustration of role taxonomy of units in networks with different generalization abilities.}
    \label{fig:unit classification}
    \end{center}
    \vskip -0.15in
\end{figure*}

However, due to the excessive capability of the DNNs, networks may be over-fitted, presenting perfect performance on $\mathcal{D}^{(train)}$ while poor performance on $\mathcal{D}^{(test)}$. These poorly-generalized networks seem to simply memorize each labels of the training samples, and their units are not able to perceive any valid feature shared in the class. In addition, \cite{DBLP:journals/corr/LiYCLH15} finds that only some of the features could be reliably learned in networks with different initialization. So, highly active units could behave quite differently between the $\mathcal{D}^{(train)}$ and the $\mathcal{D}^{(test)}$. In this case, the units that are highly active on $\mathcal{D}^{(train)}$ may be able to perceive some common features to some extent. But meanwhile, they may also be trained to perceive the features that \textbf{only} belong to $\mathcal{D}^{(train)}$ not $\mathcal{D}^{(test)}$. Since these features are not the intrinsic features in the class, the corresponding units could cause severe interference to the perception of $\mathcal{D}^{(test)}$.

Apparently, different units can present different roles on $\mathcal{D}^{(train)}$ and $\mathcal{D}^{(test)}$. Based on the functional discrepancy between $\mathcal{D}^{(train)}$ and $\mathcal{D}^{(test)}$, units can be categorized into 4 roles: core units, overfitted units, generalizing units and confusing units, as illustrated in Fig.\ref{fig:unit classification}.

\paragraph{Core units} Core units are the units that perform well on both $\mathcal{D}^{(train)}$ and $\mathcal{D}^{(test)}$,
\begin{equation}
\mU_{c} = \mU_0^{(train)} \cap \mU_0^{(test)}
\end{equation}
Typically, core units are the cornerstone units in the layer, which is colored light red in Fig. \ref{fig:unit classification}. They are usually sparse and could perceive the common features in this class for both the trained data and unseen data, so deactivating core units would cause the performance deteriorates seriously on both the two dataset. For well-generalized networks, activating only a small amount of most active units could be suffice to promote the network performance to the peak, which can be even higher than the original model. Then, the performance declines gradually along with the addition of other units, which clearly shows the difference on functionality between core units and others. For over-fitted networks, although activating the units other than core units could improve the performance on the training set, yet it could be found that these units could decrease the testing performance rapidly. 


\paragraph{Overfitted units} Overfitted units are the units being good at $\mathcal{D}^{(train)}$ but bad at $\mathcal{D}^{(test)}$,
\begin{equation}
\mU_{o} = \mU_0^{(train)} \cap (\mU_l \setminus \mU_0^{(test)})
\end{equation}
Overfitted units come after core units in the ordered list, colored orange in Fig. \ref{fig:unit classification}. The performance on training set keeps being improved, but its turning point appears on the test set obviously when overfitted units enter the network. This indicates that the functional status of these units has changed, from the previous positive effect on the network to a "false" positive effect. These units can only perceive features in the training samples. However, these features are not common features and do not exist in the test samples. Thus, overfitted units are harmful to the network because they offset the effect of core units. The existence of overfitted units implies the continuous rising of the training performance. It is a significant sign of overfitting. Remarkably, this provides a criterion to detect overfitting using only training sets.

\paragraph{Generalizing units} Generalizing units are the units that play well on $\mathcal{D}^{(test)}$ but not on $\mathcal{D}^{(train)}$,
\begin{equation}
\mU_{g} = \mU_0^{(test)} \cap (\mU_l \setminus \mU_0^{(train)})
\end{equation}
Just like overfitted units, they come after core units in the important rank, colored light yellow in Fig. \ref{fig:unit classification}. Generalizing units are the separatrix of core units and confusing units. Normally, they are rare in the network because test sets are not involved in training. Emergence of generalizing units generally implies the model is well trained. 

\paragraph{Confusing units} Confusing units are the units important on neither $\mathcal{D}^{(test)}$ nor $\mathcal{D}^{(train)}$, but may be core units to other classes,
\begin{equation}
\mU_{f} = (\mU_l \setminus \mU_0^{(train)}) \cap (\mU_l \setminus \mU_0^{(test)})
\end{equation}
They normally lie on the end part of the importance rank, colored light blue in Fig. \ref{fig:unit classification}. Confusing units are the products of performance balancing on diverse classes. Opposite to the overfitted units, if we observe the continuous slight decline in the training performance along with the addition of units, the model probably might not suffers from seriously overfitting.

Additionally, it is apparent that
\begin{equation}
    \begin{split}
        \mU_{c} \cap \mU_{o} \cap \mU_{g} \cap \mU_{f} & = \emptyset \\
        \mU_{c} \cup \mU_{o} \cup \mU_{g} \cup \mU_{f} & = \mU_l
    \end{split}
\end{equation}
\noindent which means that all units have one and only one role in the networks.

\section{Experimental Analysis of Different Unit Roles}

In our experiment, we train $5$ VGG16 models on ImageNet dataset, each of which applies a separate hyper-parameters to reach at different generalization ability. Table \ref{sample-table} shows the performance of models ranking in the descending order of the generalization gap. In general, generalization gap denotes the difference between the training accuracy and the testing accuracy. It is usually used as the indicator of the generalization ability of a DNN.

\begin{figure}
    \centering
    \begin{picture}(1.0\columnwidth, 10)
        \put(59, 0){\scalebox{0.9}{\textbf{Model A}}}
        \put(188, 0){\scalebox{0.9}{\textbf{Model B}}}
        \put(317, 0){\scalebox{0.9}{\textbf{Model C}}} 
    \end{picture}
    \vskip 0.02in
    \begin{subfigure}[t]{0.3\textwidth}
        \centering
        \includegraphics[width=1\textwidth]{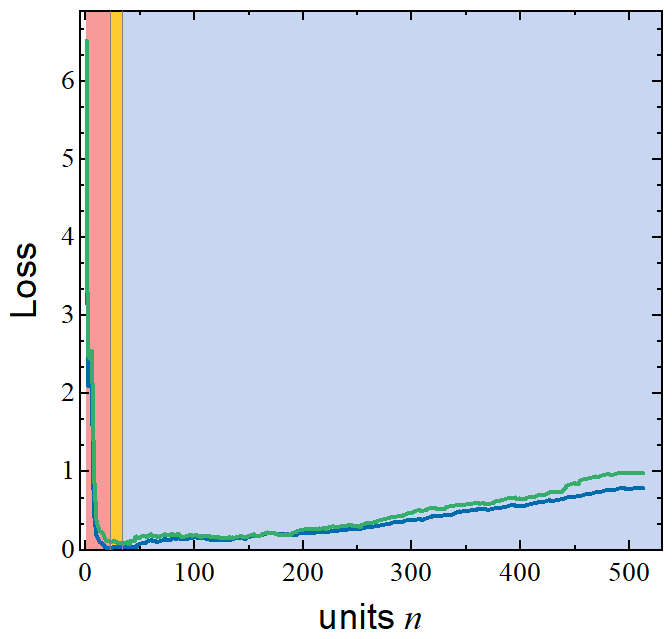}
    \end{subfigure}
    \hspace{0.05in}
    \begin{subfigure}[t]{0.3\textwidth}
        \centering
        \includegraphics[width=1\textwidth]{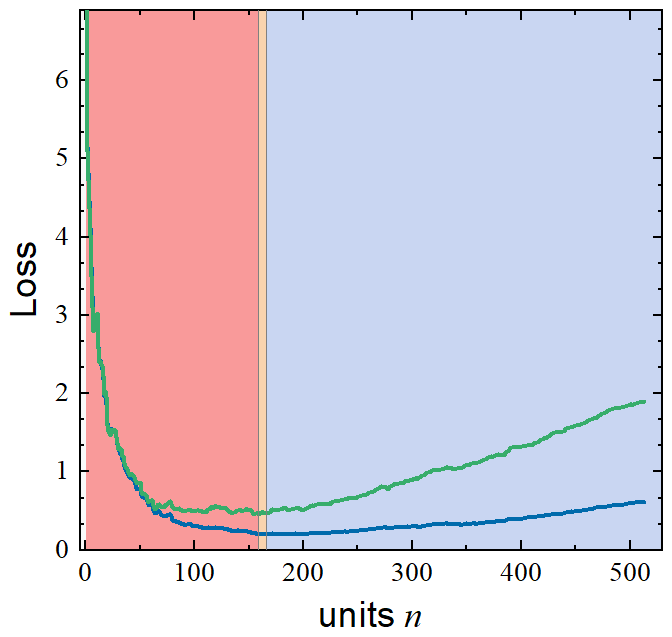}
    \end{subfigure}
    \hspace{0.05in}
    \begin{subfigure}[t]{0.3\textwidth}
        \centering
        \includegraphics[width=1\textwidth]{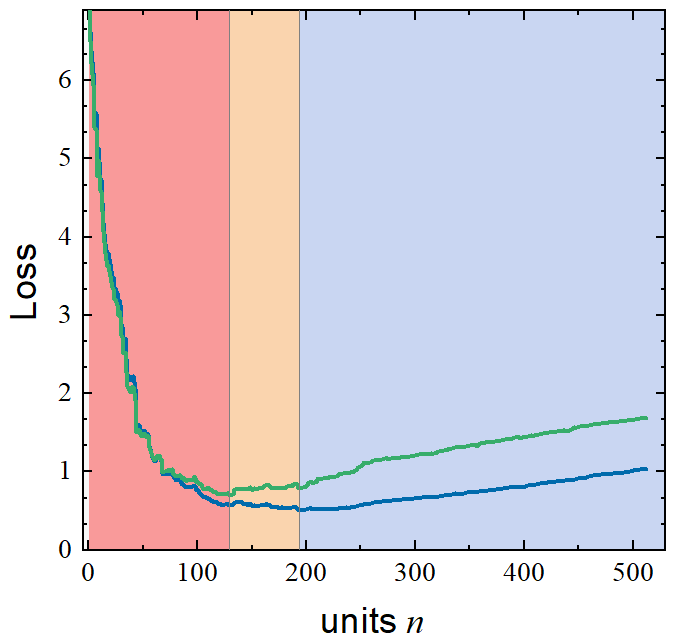}
    \end{subfigure}\\
    \begin{picture}(1.0\columnwidth, 10)
        \put(127, 0){\scalebox{0.9}{\textbf{Model D}}}
        \put(251, 0){\scalebox{0.9}{\textbf{Model E}}}
    \end{picture}
    \vskip 0.02in
    \begin{subfigure}[t]{0.3\textwidth}
        \centering
        \includegraphics[width=1\textwidth]{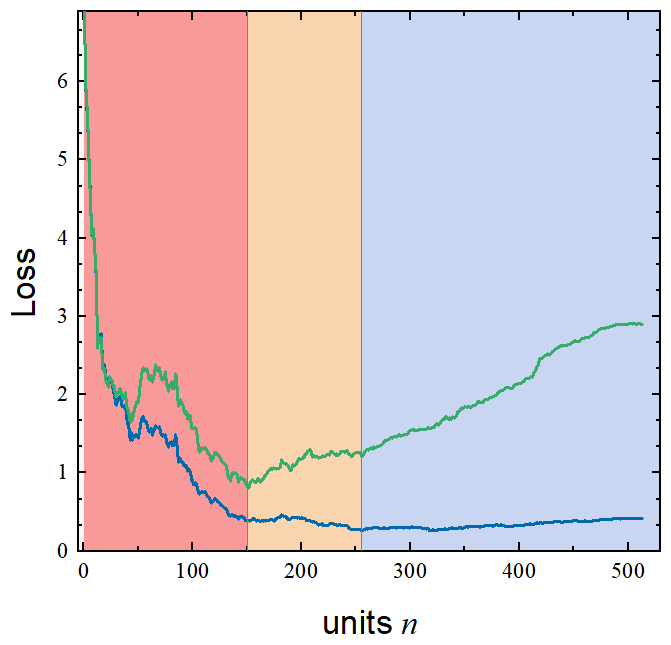}
    \end{subfigure}
    \hspace{0.05in}
    \begin{subfigure}[t]{0.3\textwidth}
        \centering
        \includegraphics[width=1\textwidth]{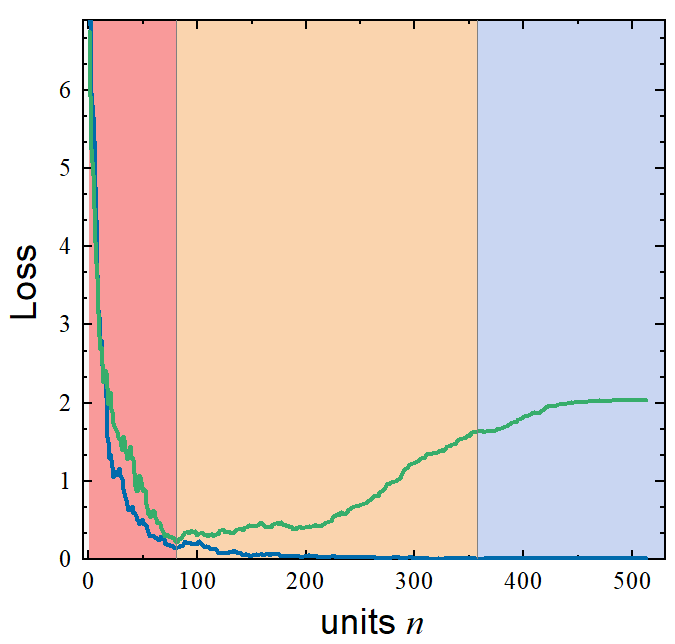}
    \end{subfigure}
    \caption{ROF curves and unit role taxonomy of $5$ VGG16 network models on the same class}
    \vskip -0.15in
    \label{c4 fig: role taxa generalization}
\end{figure}

\begin{table}[ht]
    \caption{Classification accuracies for five different models.}
    \vskip -0.05in
    \label{sample-table}
    \begin{center}
    \begin{small}
    \begin{sc}
    \begin{tabular}{lccc}
    \toprule
    Models & Training Acc & Testing Acc & Generalization Gap \\
    \midrule
    Model A & 0.732 & 0.657 & 0.075 \\
    Model B & 0.730 & 0.600 & 0.130 \\
    Model C & 0.818 & 0.543 & 0.275 \\
    Model D & 0.828 & 0.444 & 0.384 \\ 
    Model E & 0.978 & 0.374 & 0.604 \\
    \bottomrule
    \end{tabular}
    \end{sc}
    \end{small}
    \end{center}
\end{table}

We would first study the unit roles in the $5$ different networks. Fig.\ref{c4 fig: role taxa generalization} shows the results of the same class for units at the last convolutional layer in the $5$ VGG16 models. From the figure, we could found that for these well-generalized models, there are similar ROF curves on the training set and the test set. But for these poorly-generalized models, there are large differences in the ROF curves of the two data sets.At the same time, it can be seen that there are less than 30 core units in Model A, but it can give the best network model performance, indicating that these units in Model A have presented highly functional specialization which could perceive the intrinsic features in the class. As the performance of the network model continues to deteriorate, the number of overfitting units in the model will gradually increase, and the number of confusing units will continue to decrease. In addition, it can also be seen that the number of core units could keep within a certain range and will not increase as the model performance deteriorates. But generally, when there are few overfitting units, fewer core units mean that the units present better functional specialization.

\begin{figure}
    \centering
    \includegraphics[width=1.0\textwidth]{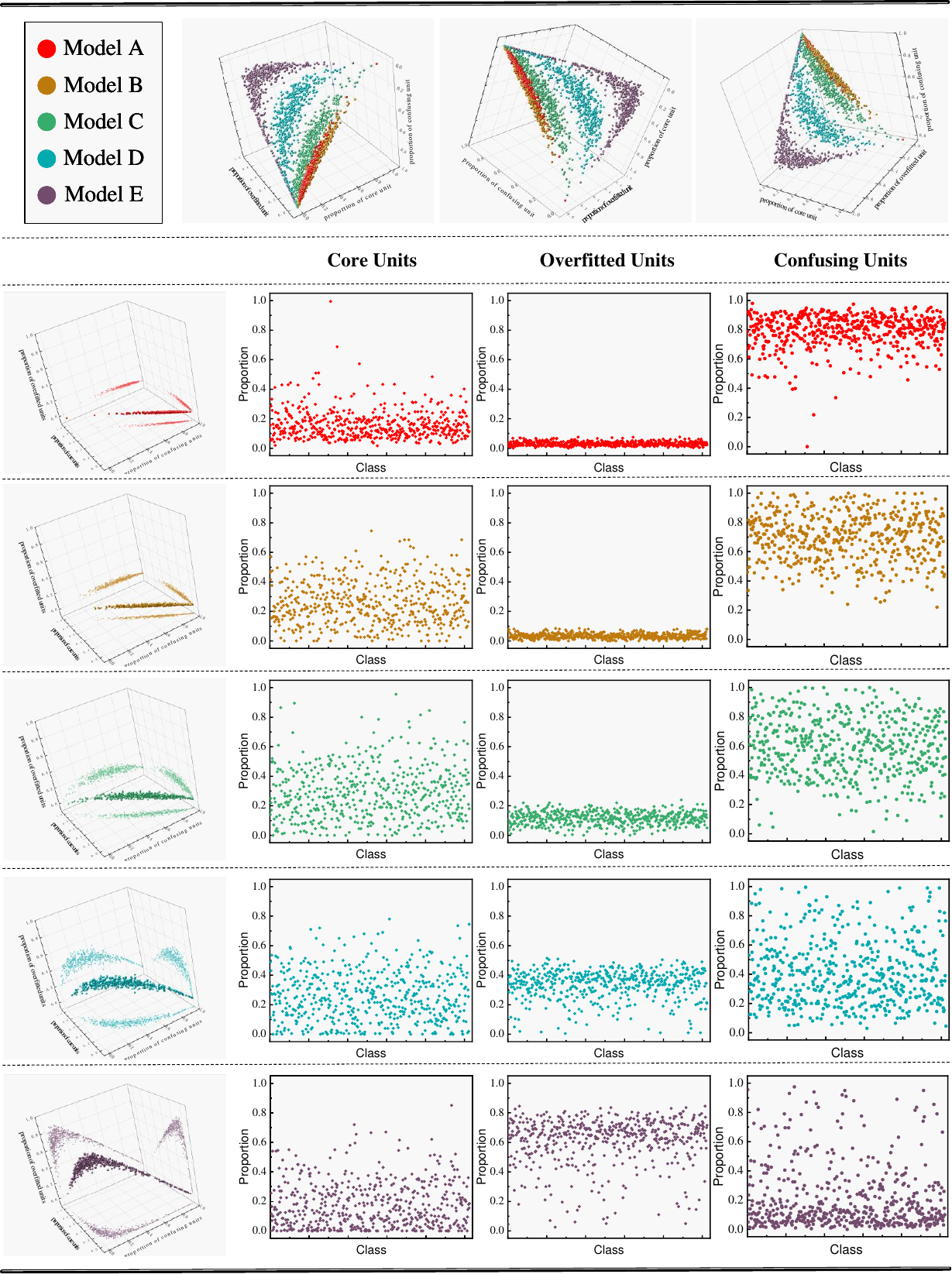}
    \vskip 0.1in
    \caption{View of the proportion of units with different roles for all classes in ImageNet}
    \label{c4 fig: unit compare}
\end{figure}

Apparently, units with different roles in network models with different performances will show different proportions. Next, the research will be extended from a single class to all classes, and the proportion of role classification of units in different classes by these models will be summarized. In order to give a clear and intuitive comparison, the proportion of unit roles corresponding to each class can be expressed as a three-dimensional vector and drawn in a three-dimensional coordinate system, as shown in \Figref{c4 fig: unit compare}. In the figure, the $x$ axis, $y$ axis and $z$ axis represent the proportion of core units, overfitted units and confusing respectively. Due to the small proportion of generalizing units, generalizing units will be ignored in the plot. Moreover, each scatter point in the figure represents the proportion of 4 different role units in a class.


It could be clearly seen from the three-dimensional comparison chart that models with different performance show well separability under the same three-dimensional coordinate system. Comparing the individual model scatter points, we can see that the role proportions between Model A and Model B are quite similar. Both models show the typical nature of well-generalized models, i.e., they have fewer core units and overfitted units, and more confusing units. In the meanwhile, it can also be seen that although Model B has the same proportion of overfitted units as Model A, Model B has more core units than Model A, and the overall variance is larger. From the previous demonstration, when the model has only a small number of overfitted units, too many core units also mean poor specialization, which will damage the performance of the network model to a certain extent. For the latter three models (C, D and E), the corresponding proportion of core units remains relatively unchanged, but as the network performance worsens, the proportion of confusing units will become smaller and the proportion of overfitted units will gradually increase. Compared with models A and B, the latter three models show completely different proportions of unit role categorization. Obviously, the role categorization ratio of units is closely related to the performance of the network.

Further, the mean and standard deviation of the unit role categorization proportions of all classes for each model will be calculated. The results are shown in table\ref{c4 table: role_taxa}. As can be seen from the table, the pattern of role proportions presented is similar to \Figref{c4 fig: unit compare}. Among them, the proportion of core units fluctuates within a range. For a well-generalized model, it should have a smaller proportion of core units and a lower standard deviation. However, the opposite is not necessarily true, such as model E. At the same time, for poorly-generalized models, they have more overfitting units. Similarly, if a network model has fewer overfitting units, it does not necessarily mean that it has a very good degree of functional specialization, such as model B. Finally, the proportion of confusing units in the model showed a strong correlation with network performance. It is worth noting that for a well-generalized model, it has a large number of units that have a negative effect on it. On the contrary, for a poorly-generalized model, there are only a small number of confusing units.

\begin{table}
    \centering
    \caption{The mean and standard deviation of the unit role categorization proportions of all classes in the $5$ VGG16 models.}
    \vskip 0.05in
    \begin{tabular}{lcccc}
    \toprule 
    Model~~~~~~~ & ~Core Units~ & Overfitted Units & Generalizing Units & ~Confusing Units~ \\
    \midrule
    Model A & $16.1\% \pm {9\%}$ & $3.1\% \pm {1\%}$ & $1.7\% \pm {1\%}$ & $79.1\% \pm {10\%}$ \\
    Model B & $25.9\% \pm {16\%}$ & $3.4\% \pm {1\%}$ & $1.8\% \pm {1\%}$ & $68.9\% \pm {17\%}$ \\   
    Model C & $28.6\% \pm {20\%}$ & $11.5\% \pm {4\%}$ & $0.5\% \pm {0\%}$ & $59.4\% \pm {22\%}$ \\   
    Model D & $25.4\%  \pm {17\%}$ & $33.9\%  \pm {11\%} $ & $0.2\%  \pm {0\%}$ & $40.5\%  \pm {24\%}$ \\    
    Model E & $17.7\% \pm {14\%}$ & $61.4\% \pm {18\%}$ & $0\% \pm {0\%}$ & $20.9\% \pm {20\%}$ \\   
    \bottomrule
    \end{tabular}
    \label{c4 table: role_taxa}
\end{table}

\section{Role Share of Individual Units}

\subsection{Role Share of Individual Units}

In this section, we would provide discussions of unit roles from another perspective. For each unit, it will play different roles when facing different classes. For example, a given unit will exist as a core unit when faced with some classes, but as a confusing unit when faced with other classes. Here, we call the proportion of roles played by neural units $\mU_n$ in all classes the \emph{Role Share},
\begin{equation}\label{eqn:roleshare1}
    RS_n=\left( RS_n^{(c)},RS_n^{(o)},RS_n^{(g)},RS_n^{(f)} \right)
 \end{equation}
where,
\begin{equation}\label{eqn:roleshare2}
    RS_n^{(\mathcal{R})} = \frac{1}{N} \sum_{k=1}^N \sI_{U_n \in \mU_{\mathcal{R}(k)}}, \ 
    \mathcal{R} = o, c, g, f 
\end{equation}
Here, $\sI(\cdot)$ denotes the indicator function.


Just as market share, role share of unit $U_k$ indicates the relative ratios of its roles (Core, Overfitted, Generalizing, Confusing) over all the data classes. The elements of $RS_n$ is not independent because
\begin{equation}\label{eqn:dependent}
   RS_n^{(c)}+RS_n^{(o)}+RS_n^{(f)}+RS_n^{(g)}=1
\end{equation}

The role share offers another dimension to delineate the specialization of network units. Cooperation and competition among units during training processes inevitably lead to the function preference of network units, which is the basis of network efficiency and generalization. The role of a unit and its share is closely associated. Table \ref{c4 table: role_share} presents the average role shares across all the neural units. It can be observed that for networks with different performance, each role share shows the same pattern as its role proportion in Table \ref{c4 table: role_taxa}. Furthermore, the average value of each role's share is highly consistent with its corresponding role proportion, implying that the neural units maintains a balanced state during training. We would give discussions from the perspective of the role share in the following.

\begin{table}
    \centering
    \caption{The mean and standard deviation of the unit role categorization proportions of all classes in the $5$ VGG16 models.}
    \vskip 0.05in
    \begin{tabular}{lcccc}
    \toprule 
    Model~~~~~~~ & ~Core Share~ & Overfitted Share & Generalizing Share & ~Confusing Share~ \\
    \midrule
    Model A & $16.1\% \pm {10\%}$ & $3.1\% \pm {1\%}$ & $1.7\% \pm {1\%}$ & $79.1\% \pm {12\%}$ \\
    Model B & $25.9\% \pm {15\%}$ & $3.4\% \pm {2\%}$ & $1.8\% \pm {1\%}$ & $68.9\% \pm {16\%}$ \\   
    Model C & $28.6\% \pm {18\%}$ & $11.5\% \pm {4\%}$ & $0.5\% \pm {0\%}$ & $59.4\% \pm {20\%}$ \\   
    Model D & $25.4\%  \pm {17\%}$ & $33.9\%  \pm {10\%} $ & $0.2\%  \pm {0\%}$ & $40.5\%  \pm {22\%}$ \\    
    Model E & $17.7\% \pm {15\%}$ & $61.4\% \pm {16\%}$ & $0\% \pm {0\%}$ & $20.9\% \pm {22\%}$ \\   
    \bottomrule
    \end{tabular}
    \label{c4 table: role_share}
\end{table}

\subsection{Discussion on Role Share}

With different generalization, trained models always present diverse distributions of units' role shares. It is interesting how should the role share be specified for a DNN with good generalization? 
 
For core unit share set $\{RS_n^{(c)}\}_{n=1}^{N}$, its element $C_n^{(c)}$ is the share of classes that the unit $U_n$ really plays a decisive part. The determination of ideal distribution for $\{RS_n^{(c)}\}_{n=1}^{N}$ is essentially a problem of multi-agent system design \cite{campbell2011multi}. In particular, one of the design objective for multi-agent system is allocating multiple tasks to multiple agents. Analogously, the unit here corresponds to the agent and the classification of images in multiple classes corresponds to the tasks. How to coordinate individual agents (units) to efficiently finish complex tasks (classification)? Typically, the agents are initialized randomly with no functionality preference. By gradually training, they automatically learn to become specialized to only a small subset of tasks \cite{murciano1997specialization}. Broad research show that specialization could reduce both the physical and virtual interference that occurs in agents and lead to overall increase in productivity of whole system \cite{campbell2011multi}. So, $RS_n^{(c)}$ should not be too large in the scenario of multiple classes for each unit ought to be in charge of only a small portion of classes. Meanwhile, when tackling the same task, several agents (units) cooperate simultaneously could improve system efficiency. Therefore, too small $RS_n^{(c)}$ should also be avoided. Hence most elements of $\{RS_n^{(c)}\}_{n=1}^{N}$ should be limited in a small neighborhood of zero, the same as that of core units (Fig\ref{c4 fig: unit compare}). The model having too many units with overly large $RS_n^{(c)}$ probably could be detriment to generalization due to the bad specialization. 

Since overfitted units are only effective on the training set, $RS_n^{(o)}$ should be as few as possible for each unit $U_n$. When a DNN is poorly generalized, large number of units will present high $RS_n^{(o)}$, meaning that for many classes, these units are overfitted, that is to say, being only effective on the training set, not on the testing set. As ablation operation being performed, removing these units will impact the training accuracy on the classes regarding these units as either core units or overfitted units. In a word, the more units with high $RS_n^{(o)}$, the more severe decline on the training accuracy in a ROF test. This results in a more dramatically drop of the overall training accuracy. This is consistent with the observations in \cite{morcos2018importance}: "training accuracy drops more rapidly for less generalized CNNs when performing cumulative ablation operation". 

However, the statement in \cite{morcos2018importance}, "a more generalized CNN relies less on single units", deserves further query. In a sense, it implies that removing a single unit will be less impact for a more generalized CNN. It seems that the difference between core and confusing units is somewhat improperly ignored. For a well generalized CNN, removing a unit will surely heavily decrease the training accuracy on the classes for which the unit is a core unit. But for many other classes where this unit exists as a confusing unit simultaneously, it will slightly increase the training performance. The total effect of removing this unit on the overall training accuracy would present less drop. That is the more comprehensive explanation for that the more generalized model seems less reliance on single units. 

For confusing unit share set $\{RS_n^{(f)}\}_{n=1}^{N}$, it should distribute around some large value close to 1, being contrary to that of core unit share set. In fact, individual confusing units in a well generalized network actually present little key contribution on the majority of classes. Our observation here somehow coincides with the actual state of famous pre-trained model BERT \cite{devlin2018bert}. Being heavily trained, BERT finally presents no task preference, providing an excellent base for solving further specific tasks with only small effort of fine-tuning. Seemingly, most units in BERT exhibit confusing functionality. Fine-tuning seeks a few units suitable for the certain task and features them to be more task-specified. That is, transforming these units from confusing units to core units. As for most unsuitable units, they keep as confusing units. It is the remarkable characteristic of well generalized models.

\section{Conclusion}

In this paper, we have provided a kind of role taxonomy for units in DNNs with different generalization ability. Given a task, units that are highly active at a layer are firstly identified by performing the introduced retrieval-of-function test. We found that they could be categorized into four types in terms of their function on separately the training set and the testing set, which are core units, overfitted units, generalizing units and confusing units. We show that the DNNs with different generalization ability would have distinct ratios of the four types of units, from both the perspective of unit role and role share. We hope our work could be helpful for further how to train a good DNN, and also expect to give connections between biological neural networks and the artificial neural networks.

\small{
\bibliographystyle{plain}
\bibliography{main}
}

\end{document}